\documentclass[11pt,twoside]{article}
\usepackage{fullpage}
\usepackage{ifthen}
\usepackage{epsf}
\usepackage{hyperref}
\usepackage{url}
\usepackage{enumerate}

 \usepackage{amsthm}
 \usepackage{amsfonts}
 \usepackage{amsmath}
 \usepackage{amssymb}
 \usepackage{color}
\usepackage{microtype}
\usepackage{multicol}
 \usepackage{amsfonts}
 \usepackage{amsmath}
 \usepackage{amssymb}

\usepackage{hyperref}
\usepackage{url}
\usepackage{amsthm}
\usepackage{amsmath}
\usepackage{amssymb}
\usepackage{algorithmic}
\usepackage{graphicx}
\usepackage[ruled,lined,english]{algorithm2e}
\theoremstyle{plain}

\theoremstyle{definition}

\theoremstyle{definition}

\providecommand{\examplename}{Example}
\providecommand{\problemname}{Problem}
\providecommand{\theoremname}{Theorem}
\usepackage{color}
\usepackage{sidecap}
\allowdisplaybreaks

\usepackage{graphics}
\usepackage{graphicx}
\usepackage{mathrsfs}
\usepackage{float}

\usepackage{epsf}
 \usepackage{subfigure}
 \usepackage{caption}
 
\newlength{\widebarargwidth}
\newlength{\widebarargheight}
\newlength{\widebarargdepth}

\makeatletter
\long\def\@makecaption#1#2{
        \vskip 0.8ex
        \setbox\@tempboxa\hbox{\small {\bf #1:} #2}
        \parindent 1.5em 
        \dimen0=\hsize
        \advance\dimen0 by -3em
        \ifdim \wd\@tempboxa >\dimen0
                \hbox to \hsize{
                        \parindent 0em
                        \hfil 
                        \parbox{\dimen0}{\def\baselinestretch{0.96}\small
                                {\bf #1.} #2
                                } 
                        \hfil}
        \else \hbox to \hsize{\hfil \box\@tempboxa \hfil}
        \fi
        }
\makeatother

\usepackage{algorithmic}
\usepackage[ruled,lined,english]{algorithm2e}

\begin{document}

\title{Learning Personalized Optimal Control for Repeatedly Operated Systems}
\author{
Theja Tulabandhula\thanks{This work was presented at the NIPS 2015 Workshop: \textit{Machine Learning From and For Adaptive User Technologies: From Active Learning \& Experimentation to Optimization \& Personalization} (ref. \url{https://sites.google.com/site/mlaihci}).} \\
Xerox Research Centre India\\
\texttt{theja@alum.mit.edu}
}
\date{December 4, 2015}
\maketitle

\begin{abstract}
We consider the problem of \emph{online learning} of optimal control for repeatedly operated systems in the presence of parametric uncertainty. During each round of operation, environment selects system parameters according to a fixed but unknown probability distribution. These parameters govern the dynamics of a plant. An agent chooses a control input to the plant and is then revealed the cost of the choice. In this setting, we design an agent that personalizes the control input to this plant taking into account the stochasticity involved. We demonstrate the effectiveness of our approach on a simulated system.

\end{abstract}

\section{Introduction}
\label{sec:introduction}

\noindent In the design of optimal control systems, one seeks a controller that performs some desired task while minimizing a given cost functional. In the classical setting, a well-defined system or plant model (i.e,. a set of differential equations governing the dynamics of the system) is assumed to be known or identified beforehand. By using this model, controllers are designed \emph{offline} by using dynamic programming (i.e., by solving the Hamilton-Jacobi-Bellman (HJB) partial differential equations) or by solving the necessary conditions provided by the Pontryagin's maximum principle (PMP) \cite{pont}. 

In this work we consider the novel problem of learning optimal controllers \emph{online} for systems that are operated repeatedly and whose models are not fully specified in each round of operation. In particular, we assume a fixed but unknown probability distribution over the parameters governing the system model. During each operation, the environment samples parameters from this distribution and they govern the system dynamics. We apply a control and are given a feedback signal on its performance at the end of operation. Over many repeated operations, our objective is to locate that control that works the best for the unknown probability distribution. This is in effect \textit{personalizing} the controller to the specific system conditions that it faces upon deployment.

Personalization has already been studied, to some effect, in the framework of online convex optimization (OCO) for the full information setting and multi-armed bandit (MAB) problems in the partial feedback setting. And these have found applications in settings such as targeted online advertisements \cite{chakrabarti2009mortal}, recommendation systems \cite{deshpande2012linear} and others. As we will see, our optimization problem is non-convex and even if full information is available, OCO cannot be applied directly \cite{zhang2015online}. On the other hand, several newer versions of bandit problems such as linear bandits \cite{filippi2010parametric}, $\chi$-armed bandits \cite{bubeck2009online} and Gaussian processes based algorithms \cite{srinivas2009gaussian} extend bandit-style algorithms to continuous domains where convexity is not always assumed. These algorithms vary in terms of what is assumed about the objective function. In terms of practicality, many of these algorithms are either too complicated for real applications or have only been shown to work on simplistic examples \cite{russo2014learning}. A closely related paper \cite{Abbasi-yadkori11regretbounds} looks at the \emph{discrete}  linear-quadratic regulator (LQR) problem where one can change the control within the operating regime. In this work, we address personalization by building on the principles behind linear bandits and develop a semi-definite programming based algorithm to assess practicality in the presence of non-convexity. 

\textbf{Applications of personalization}: Several systems such as the traffic control systems, mass transit systems, cooling systems deployed at public places, etc are repeatedly operated. They also have this feature that the system dynamics differ from one round of operation to another. For instance, the traffic profile at a junction varies from cycle to cycle. It also varies from junction to junction. An optimal controller in this setting should ideally personalize to the traffic distribution seen at its junction as well as take into account the variation in the realization of the traffic patterns at its junction. In mass transit systems such as buses and trains, the number of commuters boarding and alighting depends on the route and timing that the bus operates in. This number affects the acceleration and deceleration profile of the transit vehicle and its fuel efficiency. Thus, an acceleration and deceleration controller for the transit vehicle should personalize its control for the vehicle's route and timing. Conditional on this, it should also take into account the variation in the commuter demand encountered on this route at those times. In cooling systems deployed at large public spaces, the cooling efficiency is determined by the number of people using the space that varies based on the space characteristics and time.  Even within a specific space and time period, a cooling controller may have to take into account the variation in the usage to increase operational efficiency.



\section{\label{sec:Problem-Formulation}Problem Formulation}

We consider a continuous-time system governed by a linear constant coefficient differential equation as follows: 
\begin{align}
\dot{z}(t)=\sum_{i=1}^{p}\omega_{i}\Big(A_{i}z(t)+B_{i}u(t)\Big),
\label{eqn:lds}
\end{align}
where $\omega_{i}\in \{0,1\}$ for $i=1,...,p$, $z:[0,t_{f}]\to\mathbb{R}^{n}$ is the state of the system and $u:[0,t_{f}]\to\mathbb{R}^m$ is the control input (here $t_f$ is the time of end of control). 
Lets say want to come up with a controller $u$ to steer the system given in Equation (\ref{eqn:lds}) from a given initial condition $z_{0}\in\mathbb{R}^{n}$ to a given final condition $z(t_{f})\in\mathbb{R}^{n}$ (for simplicity, let $z(t_f)=\mathbf{0}$) minimizing a scalar valued cost given by:
\begin{align}
J(u)=\int_{0}^{t_{f}}\Big(z(t)^{T}Qz(t)+u(t)^{T}Ru(t)\Big)dt. \label{eqn:cost}
\end{align}

We assume we know the functional form of $J$ as well as matrices $Q \succ 0$ and $R \succ 0$. Without loss of generality, we can restrict our optimal control search to the space of linear feedback controls, i.e., controls of the form $u(t)=Kz(t)$, where $K$ is a matrix of gain parameters (this assumes that $t_f$ is large enough for the dynamics to settle down). Further, the gain matrix $K$ should be such that the system is stable \cite{Abbasi-yadkori11regretbounds} (we will automatically ensure this in our algorithm below).

Lets now assume that we are operating the system repeatedly. That is, we assume matrices $A_{i}\in\mathbb{R}^{n\times n}$ and $B_{i}\in\mathbb{R}^{n\times m}$ for $i=1,...,p$ are known and fixed beforehand (we also assume suitable observability and detectability conditions involving $\{A_i,B_i\}$ and $Q$). What we are not explicitly given in each round are the values of the parameters $\omega_i$, $i=1,...,p$. We assume that each $\omega_i$ is an indicator function of the event $\{\omega = i\}$ that happens with probability $\theta_i$ (thus, $\omega$ is a categorical random variable). In this setting, we are interested in searching for a controller that minimizes the expected cumulative cost of operation over all rounds. To recap, let $\mathcal{A}$ be a candidate algorithm. In each round $t$ of operation the following events occur:
\begin{itemize}
\item The environment draws a realization $\omega_t$ of $\omega$ from an unknown but fixed probability distribution $\theta$ ($\theta$ lies in a $p-1$ simplex in $\mathbb{R}^p$). The realization is kept fixed for the round. 
\item  Algorithm $\mathcal{A}$ picks a control (parametrized by $K_t$) from a set of stabilizing controllers (say $\mathcal{C}$) and applies it on the system.
\item A scalar cost value $J(K_t,\omega_t)$ is revealed to the algorithm that summarizes the cost of operation in the round.
\end{itemize}
 
The evolution of state $z(t)$ in each round of operation depends on the realization of random variable $\omega$ and the control input parametrized by $K_t$ that $\mathcal{A}$ chooses.  The expected cost of choosing a controller parametrized by $K $ is given by the map $K \mapsto\mathbb{E}_{\theta}[J(K,\omega)]$. The random variable $\omega$ models the stochasticity present in each operational cycle. For instance, the load on an autonomous vehicle/elevator changes as a function of the number of passengers alighting and boarding during each run. The number of people present at a location at various points of time also changes the loading on the corresponding cooling system in place. In the next section, we describe a solution technique that optimizes for the cumulative cost while choosing controllers.


\section{Solution Approach}
\label{sec:Proposed-Methods}

Our algorithm chooses to apply a control $K_t$ and gets to see a realization of the cost of operation $J(K_t,\omega_t)$ in each round $t$. It is able to use this feedback to deduce which realization of $\omega$ occurred\footnote{Although, such knowledge leads us to the full information setting, OCO is not applicable due to non-convexity.}. This lets it update its belief about the unknown $\theta$. Based on this belief, it \emph{optimistically} picks the next control $K_{t+1}$ to be applied. The algorithm is described in Algorithm \ref{alg:alg}. The choice of controller depends indirectly on the performances of previously explored controllers similar to many previous works \cite{Abbasi-yadkori11regretbounds,bittanti2006adaptive,hespanha2003overcoming,abbasi2014bayesian}.

\begin{algorithm}[H]
\begin{algorithmic}[1] 
\STATE Input: $\mathcal{C},c_0,\delta > 0, Q, R$
\FOR{$t=1,...$} 
\STATE{Choose $K_{t}=\arg\min_{\theta \in \Theta(c_{t-1},t,\delta)}\min_{K\in \mathcal{C}}\mathbb{E}_{\theta}[J(K,\omega)]$. } 
\STATE{ Apply $K_t$ and get feedback $J(K_t,\omega_t)$.} 
\STATE{Identify the realization $\omega_t$ of random variable $\omega$ as $i$. 
\STATE Increment the $i$-th coordinate of $c_{t-1}$ by one to get $c_{t}$.}
\ENDFOR 
\end{algorithmic}
\protect\caption{\label{alg:alg} Personalization algorithm for repeatedly operated systems}
\end{algorithm}

\textbf{Inputs}: Before Algorithm \ref{alg:alg} is deployed, we explore the system by applying different controllers for some initial set of rounds $T_{\textrm{init}}$. This gives us the initial count vector $c_0 \in \mathbb{Z}_{+}^p$ of the realizations of $\omega$ (identification of the realization is described below). In addition  to $c_0$, our algorithm also takes in a confidence parameter $\delta >0$ to be used for optimistic controller selection, the objective $J(\cdot,\cdot)$ parameterized by $Q$ and $R$ matrices and the set of stable controllers $\mathcal{C}$.

\textbf{Optimistic controller selection}: In any round $t$, we have an empirical estimate of $\theta \in \mathbb{R}_{+}^{p}$, denoted by $\hat{\theta} = \frac{c_{t-1}}{\sum_{i=1}^{p}c_{t-1}^i}$. This is similar to the maximum likelihood estimation step in linear stochastic bandits \cite{Abbasi-yadkori11regretbounds,dani2008stochastic}. By using the method of types and Pinsker's inequality (for instance, see Theorem 11.2.1 in \cite{cover2012elements}), we can upper bound the probability that the unknown $\theta$ is far from estimate $\hat{\theta}$ as:
\begin{align*}
\mathbb{P}(\|\theta - \hat{\theta}\|_{1} \geq \alpha) \leq (\tau+1)^{p}2^{\left(-\frac{\tau\alpha^2}{2}\right)},
\end{align*}
where $\tau = T_{\textrm{init}}+t-1$. If we now want to ensure that this probability is upper bounded by a value $\delta >0$, then $\theta$ belongs to the set $\{\theta: \|\theta - \hat{\theta}\|_{1} \leq \sqrt{\frac{2}{\tau}\log_{2}\left(\frac{(\tau+1)^p}{\delta}\right)}\}$ with probability at least $1 - \delta$. We define this set as $\Theta(c_{t-1},t,\delta)$. Thus while picking the controller $K_t$ for round $t$, we can optimistically search for a $\theta$ value from $\Theta(c_{t-1},t,\delta)$ simultaneously. The optimization problem (line 3 in Algorithm \ref{alg:alg}) can be written explicitly as\footnote{This formulation builds on an SDP based formulation for a deterministic LQR problem.}:
\begin{align*}
\max_{\{Y_i,L_i,\theta_i\}_{i=1}^{p}} &\sum_{i=1}^{p}\theta_i \textrm{tr}(Y_i) \;\; \textrm{subject to}&\\
\begin{bmatrix}
-(A_iY_i +B_iL_i)^T - (A_iY_i +B_iL_i)& Y_i & L_i^T\\Y_i & Q^{-1} & 0 \\ L_i & 0 & R^{-1}
\end{bmatrix} &\succ 0
\;\;\; i=1,...,p\\
L_iY_i^{-1} &= L_jY_j^{-1} \;\;\; \forall i\neq j\\
Y_i &\succ 0\;\;\; i=1,...,p\\
\|\theta - \hat{\theta}\|_{1} &\leq  \sqrt{\frac{2}{\tau}\log_{2}\left(\frac{(\tau+1)^p}{\delta}\right)}\\
\theta_i &\geq 0, \;\;\; i=1,...,p, \;\;\textrm{ and }\\
\sum_{i=1}^{p}\theta_i &= 1.
\end{align*}
This optimization problem is \emph{non-convex} and we devise some heuristics (alternating minimization over $\theta$ and $\{Y_i,L_i\}, i=1,...,p$ and a way to deal with $L_iY_i^{-1} = L_jY_j^{-1}$ coupling constraints) in the experiments. Ideally, the control $K_t$ is given by $L_iY_i^{-1}$ for any $i$.

\textbf{Identification of the realization}: In our setting, we have $p$ possible realizations of the system $\{A_i,B_i\}, i=1,...,p$. If we are given feedback $J(K_t,\omega_t)$ in round $t$, we can solve the following optimization problem with each of the $p$ pairs and the fixed control $K_t$ to get $p$ cost values $J_i(K_t), i=1,...,p$ and deduce the realization $\omega_t$:
\begin{align*}
J_i(K_t) = \min_{P,K} \textrm{tr}(P) & \;\;\;\textrm{ subject to}\\
(A_i+B_iK_t)^TP + P(A_i+B_iK_t) &+ Q + K_t^TRK_t \prec 0\\
P &\succ 0.
\end{align*}
Realization $\omega_t$ is equal to $\arg\min_{i=1,...,p} |J(K_t,\omega_t) - J_i(K_t)|$ (ties broken arbitrarily). The above non-convex optimization problem can be transformed into a semi-definite program and solved relatively easily when compared to the optimistic optimization problem for control selection formulated earlier. 

\textbf{An experts based alternative}: An alternative algorithm that is intuitive but suboptimal is as follows. We can compute the optimal controllers $K_i^*$ corresponding to each system model $\{A_i,B_i\}, i=1,...,p$ beforehand. We can then treat each of these as experts and apply the randomized weighted majority algorithm \cite{arora2012multiplicative}. We can do this because we can get full information in each round and not just the cost of the controller we picked. But note that the regret bound does not hold because the optimal controller need not belong to the set of experts $\{K_i^*\}_{i=1}^{p}$. We want to find a controller, not necessarily optimal for any of the system models $\{A_i,B_i\},i=1,...,p$, that minimizes the \emph{expected} cost of operation over multiple rounds while minimizing regret. This is what is achieved by Algorithm \ref{alg:alg}.


\section{Experiments}
\label{sec:Experiments}
We show the effectiveness of our solution approach through a three dimensional system given by
\[
\dot{z}(t)=\omega_{1}(A_{1}z(t)+Bu(t))+\omega_2(A_2z(t)+Bu(t)),
\]
with $A_1=\begin{bmatrix}0 & 1 & -1\\ 0 & 0 & 1 \\ 0 & 0 & 0\end{bmatrix}$,
$A_2=\begin{bmatrix}0 & 1 & 1\\ 0 & 0 & 1 \\ 0 & 0 & 0\end{bmatrix}$, 
$B=\begin{bmatrix} 0 \\ 1\\ 1\end{bmatrix}$ 
and $u(t)=Kz(t)$. 
Further, $\omega_1 = \mathbf{1}_{[\omega = 1]}$ and $\omega_2 = \mathbf{1}_{[\omega = 2]}$,  where $\omega$ is a categorical random variable with a fixed probability mass function (unknown to the algorithm) given by $\theta = [0.5 \;\;0.5]$. Further, we set $Q =\begin{bmatrix}1 & 0 & 0\\ 0 & 1 & 0 \\ 0 & 0 & 1\end{bmatrix} $ and $R = [1\;\; 1\;\; 1]$. 

\begin{SCfigure}
  \centering
  \caption{Total cost of operation (lower is better) over 30 rounds for various controller selection schemes for personalizing to the given 3-dimensional dynamical system. Our controller scheme (Kproposed) performs the best. Although the optimal controller K2 is performing as well as Kproposed, it is not known a priori.}
  \includegraphics[width=0.45\textwidth]{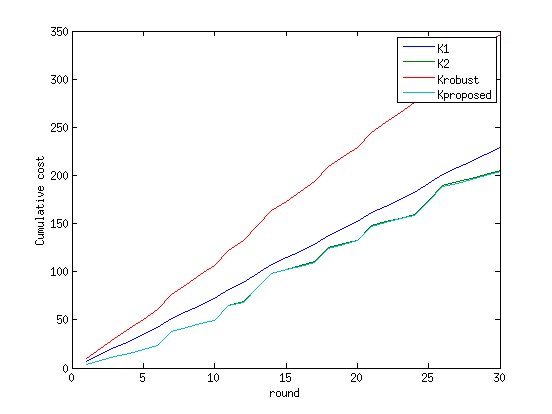}
\label{fig:experiment}
\end{SCfigure}

The experiment is run for 30 rounds and in each round, a controller (Kproposed) is chosen according to Algorithm \ref{alg:alg} and the cost it incurs is logged. We also evaluate the following static controllers: (K1) the optimal controller for $\{A_1,B\}$, (K2) optimal controller for $\{A_2,B\}$, and (Krobust) the robust optimal controller. The performances of all these controllers are plotted in Figure \ref{fig:experiment}. We observe that the proposed controller is the best in terms of the total cost accumulated. Notice that controller K2 also accumulates similar cumulative cost, and is a good choice as well, but this is not known a priori to a learning agent.

\section{Conclusions and future directions}
\label{sec:Conclusions}

In this work, we proposed an approach to personalize control systems to the operating environment in the setting where there is repetition of operation and a certain type of stochasticity is present. In particular, we proposed an algorithm that uses the \emph{optimism under uncertainty} principle. This way of personalization is very useful in reducing operational costs in a variety of applications (for instance, minimizing energy consumption in various transportation and cooling system applications).

This is still a work in progress and investigating bounds on regret for this setting is of immediate interest. It is also interesting to explore better algorithms to deal with the non-convex optimization problem that needs to be solved in each round. Also, the uncertainty model can be extended to the setting where there is a Dirichlet prior on the the unknown probability distribution $\theta$. Extensions to classes of non-linear and noisy dynamical systems are also worth pursuing.

\section*{Acknowledgement}
The author would like to thank Deepak Patil for initial discussions on this topic.

\bibliographystyle{myIEEEtran}
\bibliography{learning}
\end{document}